\documentclass[11pt,letterpaper]{article}
\usepackage{naaclhlt2009}
\usepackage[authoryear,round]{natbib}
\usepackage{times}
\usepackage{latexsym}
\usepackage{xspace}
\usepackage{graphicx}
\usepackage[breaklinks=true]{hyperref}
\usepackage{breakurl}
\usepackage[dvipsnames]{color}
\usepackage{multicol}
\usepackage[overload]{textcase}
\usepackage{setspace}

\newcommand{\ex}[1]{`{\it #1}'}

\newcommand{\figtext}[1]{#1}

\newcommand{\dm}{DE\xspace}
\newcommand{\nddm}{DE(ND)}
\newcommand{\nddmlong}{\dmlong (narrowly-defined)\xspace}
\newcommand{\monotonic}{entailing}
\newcommand{\dmlong}{downward \monotonic\xspace}
\newcommand{\dmlonghyph}{downward-\monotonic\xspace}
\newcommand{\Dmlonghyph}{Downward-\monotonic\xspace}
\newcommand{\dmnp}{downward entailment\xspace}
\newcommand{\umlong}{upward \monotonic\xspace}
\newcommand{\oplong}{operator\xspace}
\newcommand{\opslong}{operators\xspace}
\newcommand{\po}{\dm \oplong}
\newcommand{\podash}{\dm-\oplong}
\newcommand{\polong}{\dmlonghyph \oplong}
\newcommand{\pos}{{\po}s\xspace}

\newcommand{\poslong}{{\polong}s\xspace}

\newcommand{\Polongs}{\Dmlonghyph\ {\oplong}s\xspace}
\newcommand{\polongs}{{\polong}s\xspace}
\newcommand{\candidate}{candidate\xspace}
\newcommand{\candidates}{candidates\xspace}
\newcommand{\plausible}{plausible\xspace}
\newcommand{\presentence}{sentence\xspace}

\newcommand{\presentences}{{\presentence}s\xspace}
\newcommand{\sentvar}{p}
\newcommand{\scorefn}{S}
\newcommand{\distscorefn}{\scorefn_d}
\newcommand{\distillation}{distillation\xspace}
\newcommand{\Distillation}{Distillation\xspace}
\newcommand{\npicontext}{NPI context\xspace}
\newcommand{\npicontexthyph}{NPI-context\xspace}
\newcommand{\npicontexts}{{\npicontext}s\xspace}
\newcommand{\marknpi}[1]{${\mbox{\it #1}}_{{\rm NPI}}$}

\newcommand{\umex}{know\xspace}
\newcommand{\dmex}{doubt\xspace}
\newcommand{\predextrarestrict}{via fleas\xspace}
\newcommand{\predorigrestrict}{quickly\xspace}
\newcommand{\predbackbone}{spread\xspace}
\newcommand{\stmt}[2]{We {#1} the epidemic #2}
\newcommand{\pred}[1]{\predbackbone #1}\newcommand{\predorig}{\pred{\predorigrestrict}}\newcommand{\predrelax}{\pred{}}\newcommand{\predrestrict}{\pred{\predorigrestrict \predextrarestrict}}\newcommand{\stmtrelax}[1]{\stmt{#1}{\predrelax}}\newcommand{\stmtrestrict}[1]{\stmt{#1}{\predrestrict}}\newcommand{\stmtorig}[1]{\stmt{#1}{\predorig}}

\newcommand{\povar}{d} 

\newcommand{\npisubscript}{{\rm byNPI}} 
\newcommand{\npisentp}{F_\npisubscript}
\newcommand{\sentp}{\npisentp}
\newcommand{\nontrivsentp}{\widehat{F}_\npisubscript}
\newcommand{\corpusp}{F}

\newcommand{\distractor}{distractor\xspace}
\newcommand{\distractors}{{\distractor}s\xspace}
\newcommand{\cvar}{c} 

\newcommand{\no}{not-\dm}

\title{Without a 
`doubt'?\\ Unsupervised discovery of \poslong}
\author{Cristian Danescu-Niculescu-Mizil, Lillian Lee, and Richard Ducott \\ Department of Computer Science \\ Cornell University \\ Ithaca, NY 14853-7501 \\ 
{cristian@cs.cornell.edu, llee@cs.cornell.edu, rad47@cornell.edu}}

\begin{document}
\providecommand{\pdfpageheight}{\paperheight}
\providecommand{\pdfpagewidth}{\paperwidth}

\setlength{\pdfpageheight}{\paperheight}
\setlength{\pdfpagewidth}{\paperwidth}

\maketitle
\begin{abstract}
An important part of
textual inference is  making
deductions involving {\em monotonicity},
that is, determining whether a given assertion entails restrictions
or relaxations of that assertion.  For instance,
the statement \ex{\stmtorig{\underline{\umex}}} does not entail 
\ex{\stmtrestrict{\umex}}, but  \ex{\stmtorig{\underline{\dmex}}}
entails 
\ex{\stmtrestrict{\dmex}}.  
Here, 
we present the first algorithm for the challenging 
lexical-semantics
problem of learning
linguistic constructions that, like \ex{doubt}, are 
\emph{\dmlong (\dm)}.
Our
algorithm is unsupervised,  resource-lean, and effective, accurately recovering
many \pos that are missing from the hand-constructed lists that
textual-inference systems currently use. 

{\it Publication venue:} NAACL HLT 2009 

\end{abstract}

\section{Introduction}

Making inferences based on natural-language statements is a crucial
part of true natural-language understanding, and thus has many
important applications.  As the field of NLP has matured, there has
been a resurgence of interest in 
creating systems capable of making such inferences, as
evidenced by the activity surrounding the ongoing sequence of 
``Recognizing Textual Entailment'' (RTE) competitions
\citep{Dagan+Glickman+Magnini:06a,Bar-Haim+al:06a,Giampiccolo+al:07a} and the AQUAINT knowledge-based evaluation project \citep{Crouch+Sauri+Fowler:05a}.

The following two examples help illustrate the 
particular
type of inference that is the focus of this paper.
\begin{enumerate} 
\item \newcounter{knowlabel}\setcounter{knowlabel}{\value{enumi}} \ex{\stmtorig{\underline{\umex}}}
\item \newcounter{doubtlabel}\setcounter{doubtlabel}{\value{enumi}} \ex{\stmtorig{\underline{\dmex}}}
\end{enumerate}
A relaxation of  \ex{\predorig} is \ex{\predrelax}; a
restriction of it is \ex{\predrestrict}.  From
statement \theknowlabel, we can infer the relaxed version,
\ex{\stmtrelax{\umex}}, whereas the restricted
version, \ex{\stmtrestrict{\umex}}, does not follow.  But the reverse holds for
statement \thedoubtlabel: it entails the restricted version
\ex{\stmtrestrict{\dmex}}, but not the relaxed version.
The reason is that 
\ex{\dmex} is a {\em \polong% 
};\footnote{Synonyms for ``\dmlong''
include {\em downward-monotonic} and {\em monotone decreasing}. 
Related concepts
include {\em anti-additivity}, {\em veridicality}, and {\em one-way implicatives}.} 
in other words, it allows one to, in a sense, ``reason from sets to
subsets'' \citep[pg. 90]{vanderWouden:97a}.

\Polongs are not restricted to assertions about belief or to verbs.
For example,  the 
preposition \ex{without} is also \dmlong: from \ex{The applicants came without payment or
  waivers} we can infer that all the applicants came without payment.
(Contrast this with \ex{with}, which, like \ex{\umex}, is {\em \umlong}.)
In fact, there are many \polongs, encompassing many syntactic types;
these
include explicit negations like \ex{no} and \ex{never}, but also many
other terms, such as 
\ex{refuse (to)}, \ex{preventing},
\ex{nothing}, 
\ex{rarely}, and
\ex{too [adjective] to}.

As the prevalence of these operators indicates and as \citet[pg. 92]{vanderWouden:97a} states, \dmnp ``plays an
extremely important role in natural language''
\citep{vanBenthem:86a,Hoeksema:86a,SanchezValencia:91a,Dowty:94a,MacCartney+Manning:07a}.
Yet to date, only a few systems attempt to handle the phenomenon
in a general way, i.e., to consider more than simple direct negation
\citep{Nairn+Condoravdi+Kartunnen:06a,MacCartney+Manning:08a,Christodoulopoulos:08a,Bar-Haim+al:08a}.  These systems
rely on 
lists
of items annotated with respect to their behavior in ``polar''
(positive or negative) environments.  
The lists contain a relatively small number 
of \dmlonghyph
operators, at least in part because they were constructed mainly by manual inspection of verb lists
(although a few non-verbs are sometimes also included).
{\em We therefore propose to automatically
learn 
{\polongs}\footnote{
We include superlatives (\ex{tallest}), comparatives (\ex{taller}), and conditionals (\ex{if}) in this category because they have non-default (i.e., non-\umlong) properties --- for instance,  \ex{he is the tallest father} does not entail \ex{he is the tallest man}.  Thus, they also require special treatment when considering entailment relations. In fact, there have been some attempts to unify these various types of non-\umlong \opslong \citep{vonFintel:99a}.   We use the term {\em \nddmlong (\nddm)} when we wish to specifically exclude superlatives, comparatives, and conditionals.
}
--- henceforth \pos for
short
--- from data; deriving more comprehensive lists of \pos in this
manner promises to
substantially enhance the ability of textual-inference systems to
handle monotonicity-related phenomena.}

\paragraph{Summary of our approach} There are a number of significant challenges to applying a
learning-based approach.  First, to our knowledge there do not exist
\dm-operator-annotated corpora, and 
moreover, relevant types of semantic information
are ``not available in or deducible from any public lexical database''
\citep{Nairn+Condoravdi+Kartunnen:06a}.  Also, 
it seems there is no simple test one can apply to all
possible candidates;
\citet[pg. 110]{vanderWouden:97a} 
remarks, ``As a rule of thumb,
assume that everything that feels negative, and everything that
[satisfies a condition described below], is monotone decreasing.  This rule of
thumb will be shown to be wrong as it stands; but it sort of works,
like any rule of thumb.''

Our first insight into how to overcome these challenges is to
leverage a finding from the linguistics literature, {\em Ladusaw's
\citeyearpar{Ladusaw:80a} hypothesis}, which can be treated as a cue
regarding the distribution of \pos: it asserts that a certain class of
lexical constructions
known as {\em negative polarity items (NPIs)} can only appear in
the scope of \pos.
Note that this hypothesis suggests that one can develop 
an {\em
unsupervised} algorithm  based simply on checking for co-occurrence with
known NPIs.

But there are significant problems with applying this
idea in practice, including: (a) there is no
agreed-upon list of NPIs;
(b) terms can be ambiguous with respect to
NPI-hood; and (c) many non-\pos tend to co-occur with NPIs as well.
To cope with these issues, we develop a novel 
unsupervised
{\em \distillation}
algorithm that helps
filter out the noise introduced by these
problems.  This algorithm is very effective: 
it is accurate and derives many \pos that do not 
appear
on pre-existing lists.

\paragraph{Contributions} 
Our project draws a connection between the creation of textual entailment systems and linguistic inquiry regarding \pos and NPIs, and thus
relates to
 both language-engineering and linguistic concerns.

To our knowledge, this work represents the
first attempt to 
aid in the process of {\em discovering} \pos, a task whose importance we have
highlighted above.  
At the very least, our method can be  used to provide high-quality raw materials to help  human annotators create more extensive \po lists.
In fact, 
while 
previous 
manual-classification efforts have
mainly
focused on verbs, we 
retrieve \pos across multiple parts of speech.
Also,
although we discover many items (including verbs) that are not on
pre-existing manually-constructed lists, the
items we find occur frequently --- they are not somehow peculiar or
rare.  

Our
algorithm is surprisingly 
accurate
given that it is quite resource- and knowledge-lean.  Specifically, it relies only on Ladusaw's hypothesis as initial inspiration,  a relatively short and arguably noisy list of NPIs, and a large unannotated corpus.  It does {\em not} use other linguistic information --- for example, we do not use parse information, even though c-command relations have been asserted to play a key role in the licensing of NPIs \citep{vanderWouden:97a}.

\section{Method}
\label{sec:method}

We mentioned in the introduction some significant challenges to
developing a machine-learning approach to discovering \pos.  The key
insight we apply to surmount these challenges is that in the
linguistics literature, it has been hypothesized that there is a
strong connection between \pos and
{\em negative polarity items (NPIs)}, which are terms
that tend to occur in ``negative environments''.  An example NPI is
\ex{anymore}: one can say \ex{We don't have those anymore} but not
\ex{$*$We have those anymore}.

Specifically, we propose to take advantage of the seminal hypothesis of \citet[influenced
by \citet{Fauconnier:75a}, inter alia]{Ladusaw:80a}:
\begin{quote}
(Ladusaw) NPIs only appear within the scope of \dmlonghyph operators. 
\end{quote}
This hypothesis has
been actively discussed, updated,  and contested by multiple parties
\citep[inter alia]{Linebarger:87a,vonFintel:99a,Giannakidou:02a}.  It is 
not
our intent to comment 
(directly)
on its overall validity.  Rather, we simply view it as a very useful 
starting point for developing
computational tools to find \pos --- indeed, even detractors of the
theory have called it ``impressively algorithmic''
\citep[pg. 361]{Linebarger:87a}.

First, a word about scope.
For Ladusaw's hypothesis, scope should arguably be defined in terms of
c-command, immediate scope, and so on \citep[pg. 100]{vonFintel:99a}. 
But for simplicity and to make our approach as resource-lean as
possible, we simply assume that
potential \pos occur 
to the left of NPIs,\footnote{There are 
a few
exceptions, such as with the NPI ``for the life of me''
\citep{Hoeksema:93a}.}
except that we ignore text to the left of any preceding commas or
semi-colons as a way to enforce a degree of locality. 
 For example,
in both
\ex{By the way, we don't have plants 
\marknpi{anymore} because they died} and
\ex{we don't have plants \marknpi{anymore}}, we
look for \pos within the sequence of words \ex{we don't have plants}.
We refer to such sequences in which we seek \pos as {\em \npicontexts}.

Now, 
Ladusaw's hypothesis suggests that we can find \pos by looking 
for words that tend to occur more often in 
\npicontexts
than 
they occur overall.
We formulate this as follows:
\begin{quote}
\textit{Assumption:} 
For any \po $\povar$, $
\npisentp(\povar)
> \corpusp(\povar)$.
\end{quote}
Here,  $\npisentp(\povar)$ is the number of occurrences of
$\povar$ in 
{\npicontexts}\footnote{
Even if  $\povar$ occurs multiple times in a single 
\npicontext we only count it once; this way we ``dampen the signal'' of function words that can
potentially occur
multiple times in a single sentence.}  divided by the number of 
words
in \npicontexts, and 
$\corpusp(x)$ refers to the number of occurrences of $x$ relative to
the number of words in the corpus.

An additional consideration is that we would like to focus on the
discovery of {\em novel} or non-obvious \pos.  Therefore, 
for a given candidate \po $\cvar$, we compute $\nontrivsentp(c)$: the value of $\sentp(\cvar)$
that results {\em if we discard all \npicontexts 
containing a \po on a list of 10 well-known instances}, 
namely, \ex{not}, \ex{n't}, \ex{no}, \ex{none}, \ex{neither}, \ex{nor}, \ex{few}, \ex{each}, \ex{every}, and \ex{without}.
(This list is based on the list of \pos used by the RTE system presented 
in \citet{MacCartney+Manning:08a}.)
This yields the following scoring function:
\begin{equation}
\scorefn(\cvar) :=  
\frac{\nontrivsentp(\cvar)}{\corpusp(\cvar)}.%, & \mbox{if} \min(\npisentp(\cvar),\corpusp(\cvar)) \geq %\threshold \\
\label{un_distilled_score}
\end{equation}

\paragraph{\Distillation}
There are certain terms that are not \pos, but 
nonetheless 
co-occur with NPIs as a side-effect of 
co-occurring with true \pos themselves.
For instance, 
the proper noun \ex{Milken}
(referring to  Michael
Milken, the so-called ``junk-bond king") occurs relatively frequently with the \po \ex{denies}, and \ex{vigorously} occurs frequently with \pos like 
\ex{deny} and \ex{oppose}.  We refer to 
terms like \ex{milken} and \ex{vigorously}
as ``piggybackers",  
and address the piggybackers problem by leveraging the following intuition: in
 general, we do not expect to have two \pos in the same \npicontext.\footnote{One reason is that if two \pos are composed, they ordinarily create a positive context, which would not license NPIs (although this is not always the case \citep{Dowty:94a}).} 
One 
way to implement this would be to re-score the candidates in a winner-takes-all fashion: for each 
\npicontext, reward only 
the candidate with the highest score $\scorefn$.
However, 
such a
method is too aggressive
because
it would force us to pick 
a single candidate
even
when there are several with relatively close scores ---
and we know our score $\scorefn$ is imperfect.
Instead, we propose the following ``soft" mechanism.
Each sentence distributes a
``budget'' of total score
1 among the
candidates it contains according to the relative scores of those
candidates; this works out to yield the following new {\em
distilled} scoring 
function
\begin{equation}
\distscorefn(\cvar)=\frac{\sum_{{\rm \npicontexts} \,\,\,\sentvar}\ \ \frac{\scorefn(c)}{n(\sentvar)}}{N(\cvar)}
,
\end{equation}
where $n(\sentvar)={\sum_{\cvar \in\,\sentvar }\scorefn(\cvar)}$ is 
an
\npicontexthyph
normalizing factor and $N(\cvar)$ is the number of \npicontexts containing the candidate $c$. 
This way, \plausible \candidates that 
have high $\scorefn$ scores 
relative to the other candidates in
the \presentence 
receive enhanced $\distscorefn$ scores.
To put it another way:
apparently \plausible \candidates that often appear in \presentences with multiple good candidates (i.e., piggybackers) receive a low distilled score, despite a high initial 
score.

Our general claim is that the higher the   distilled score of a
\candidate, the better its chances of being a \po.

\paragraph{Choice of NPIs}  
Our proposed method requires access to a set of NPIs.  However, there
does not appear to be universal agreement on such a set.
\citet{Lichte+Soehn:07a} mention some doubts regarding approximately
200 (!) 
of the
items on a roughly 350-item list of German NPIs
\citep{Kuerschner:83a}.  
For English, the ``moderately
complete''\footnote{\url{www-personal.umich.edu/~jlawler/aue/npi.html}}
\citet{Lawler:05a} list contains two to three dozen items;  however,
there is also a list of English NPIs that is several times longer
\citep[written in German]{vonBergen+vonBergen:93a},
and 
\citet{Hoeksema:97a} asserts that English should have
hundreds of NPIs, similarly to French and Dutch.  

We 
choose to focus on the items on these lists that seem most
likely to be effective cues for our task.  
Specifically, we 
select a subset of the Lawler NPIs, focusing mostly
on those that do not have a 
relatively frequent
non-NPI sense.
An example discard is
\ex{much}, whose NPI-hood depends on what it modifies and perhaps
on whether there are degree adverbs pre-modifying it
\citep{Hoeksema:97a}.
There are some ambiguous NPIs that we do retain due to their frequency.
For example, \ex{any} 
occurs both in a non-NPI
``free
choice'' variant, as in \ex{any idiot can do that}, and 
in
an NPI version.  Although it is ambiguous with respect to NPI-hood, 
\ex{any} is also a very valuable cue due to its frequency.%
\footnote{It is by far the most frequent NPI, appearing in
36,554
of the sentences in the BLLIP corpus (see Section \ref{sec:results}).
}
Here is our NPI list:
{\footnotesize
\begin{tabular}{|l|l|l|l|} \hline
any &  in weeks/ages/years & budge & yet\\
at all & drink a drop & red cent & ever \\
 give a damn & last/be/take long &  but what & bother to \\
do a thing & arrive/leave 
until & give a shit & lift a finger \\
bat an eye & would care/mind & eat a bite & to speak of   \\\hline
\end{tabular}
}

\section{Experiments}\label{sec:results}

Our main set of evaluations focuses
on the precision 
 of our method
at discovering new \pos.  We then briefly discuss evaluation of other dimensions.

\newcommand{\bargraphcaptiontext}[2]{
(a) Precision at $k$ for $k$ 
divisible by 10 up to $k= 150$.  The bar divisions are, from the x-axis up, \nddm\ (blue, the largest); 
Superlatives/Conditionals/Comparatives (green, 2nd largest); and Hard (red, sometimes non-existent).
#2.
}

\subsection{Setup}

We applied our method to the entirety  of the 
BLLIP (Brown Laboratory for Linguistic Information Processing)
1987--89 WSJ Corpus Release 1, available from the LDC (LDC2000T43).
The 1,796,379 sentences in the 
corpus
comprise
53,064
\npicontexts;
after discarding the ones 
containing the 10 well-known \pos,
30,889 \npicontexts 
were
left.  
To avoid sparse data problems, we did not consider \candidates 
with
very low frequency in the corpus ($\leq$150 occurrences) or in the \npicontexts ($\leq$10 occurrences).

\paragraph{Methodology for eliciting judgments}

The obvious way to evaluate the precision of our algorithm is to have
human annotators judge each output item as to whether it is a \po or
not. However, there are some methodological issues that arise.  

First, if the judges know that every term they are rating comes from
our system and that we are hoping that the algorithm extracts \pos,
they may be biased towards calling every item ``\dm'' regardless of
whether it actually is.  We deal with this problem by introducing
{\em \distractors} --- items that are not produced by our algorithm,
but are similar enough to not be easily identifiable as
``fakes''.    Specifically, for each possible part of speech of
each of our system's outputs $\cvar$ that
exists in WordNet, we choose a \distractor that is either in a
``sibling'' synset (a hyponym of $\cvar$'s hypernym) or an antonym.
Thus, the \distractors are highly related to the candidates.  Note that they
may in fact also be \pos.

The judges were made aware of the presence of a substantial number of
\distractors (about 70 for the set of top 150 outputs). This design
choice did seem to help ensure that the judges carefully evaluated
each item. 

The second issue is that, as mentioned in the introduction, there does not seem to be a
uniform test that judges can apply to all items to ascertain their
\dm-ness; but we do not want the judges to  improvise excessively,
since that can introduce undesirable
randomness into their decisions.  We therefore encouraged the judges
to try to construct sentences wherein the arguments for candidate \pos
were drawn from a set of
phrases and restricted replacements we specified 
(example: \ex{singing} vs \ex{singing loudly}).
However, improvisation was still required in a number of cases; for
example, the 
candidate \ex{act}, as either a noun or
a verb, cannot take \ex{singing} as an argument.

The labels that the judges could apply were 
``\nddm'' (\nddmlong), ``superlative'',
``comparative'', ``conditional'', ``hard to tell'', and 
``\no'' (= none
of the above).  We chose this fine-grained sub-division because the
second through fourth categories are all known to co-occur with NPIs.
There
is some debate in the linguistics literature as to whether they
can be considered to be 
\dmlong, narrowly construed,  or not \citep[inter alia]{vonFintel:99a}, but nonetheless,
such operators call for special reasoning quite distinct from that
required when dealing with \umlong operators --- hence, we consider it
a success when our algorithm identifies them.

Since monotonicity phenomena can be rather subtle, the judges engaged
in a collaborative process.  Judge A 
(the second author)
annotated all 
items, but
worked in batches of around 10 
items.  At the end
of each 
batch, 
Judge B 
(the first author)
reviewed Judge A's decisions,
and the two consulted to resolve disagreements
as far as possible.

One final remark regarding the annotation: some decisions still seem
uncertain,
since various factors such
as context, Gricean maxims, what should be presupposed\footnote{For example, \ex{X doubts the epidemic spread quickly} might be said to entail \ex{X would doubt the epidemic spreads via fleas, presupposing that X thinks about the flea issue}.} and so on come
into play.  However, we take comfort in a comment by Eugene Charniak
(personal communication) to the effect that if a word causes a native
speaker to pause, that word is interesting enough to be included.  And
indeed, it seems reasonable that if a native speaker thinks there
might be a sense in which a word can be considered \dmlong, then our
system should flag it as a word that an RTE system should at least perhaps pass to a different subsystem for further analysis.

\subsection{Precision Results}

\begin{figure*}[htp]
\begin{center}
\begin{small}
\begin{tabular}{cc}
\includegraphics[width=3in,viewport= 50 220 550 557,clip]{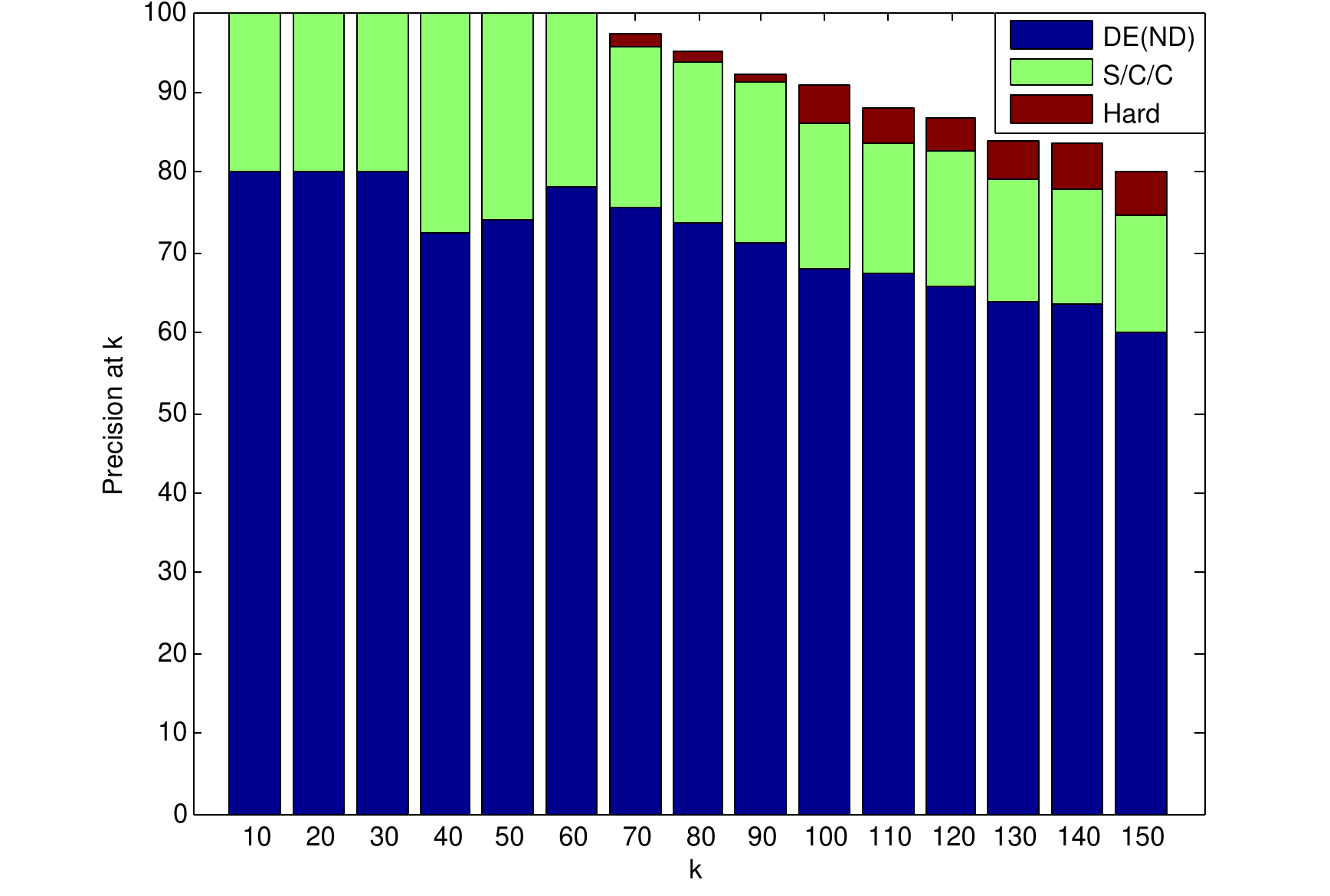}
&\includegraphics[width=3in,viewport= 50 220 550 557,clip]{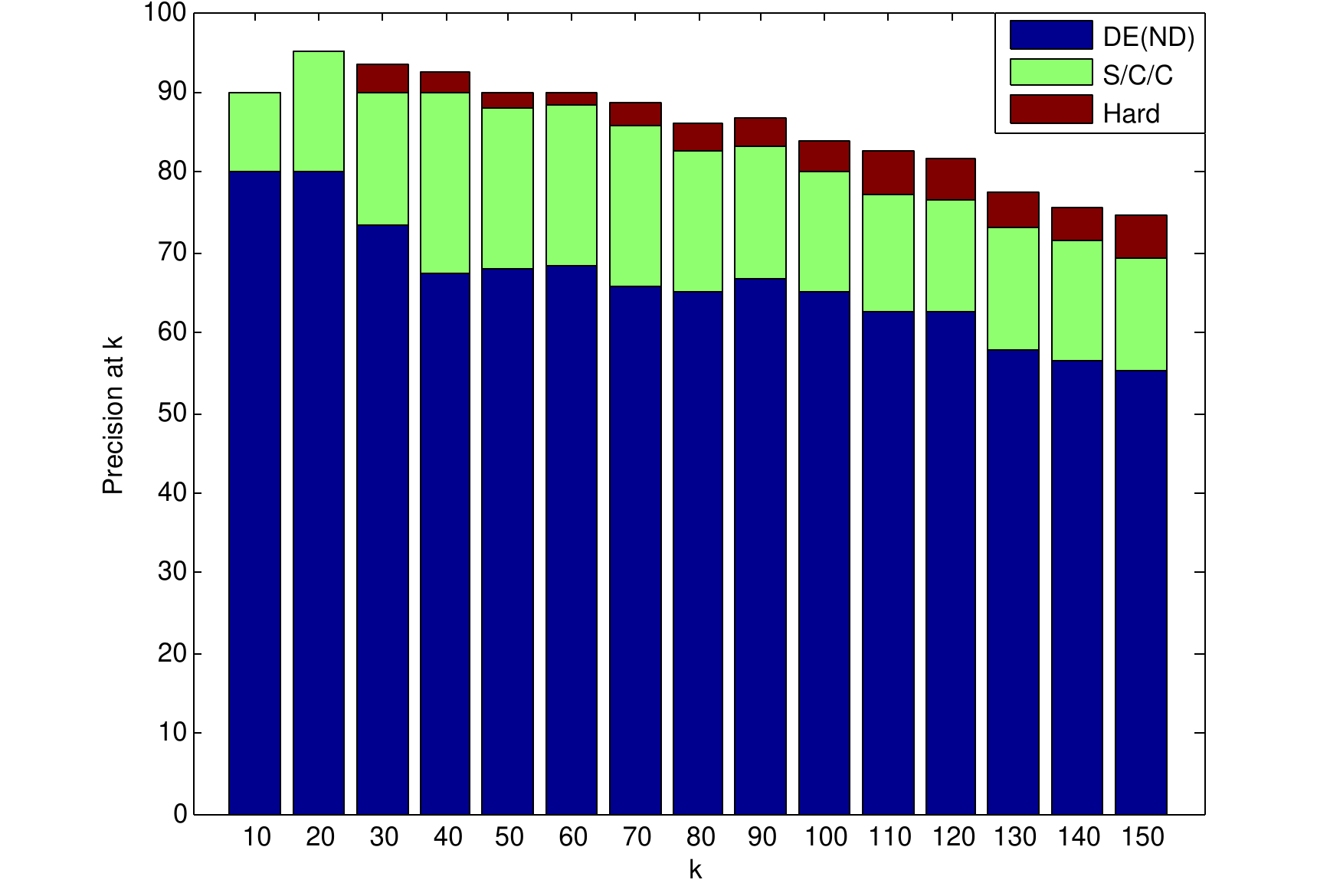}\\
(a)&(b)
\end{tabular}
\end{small}
\end{center}
\caption{\label{fig:bars} \bargraphcaptiontext{}{For example, all of the first 10 outputs were judged to be either  \nddmlong (8 of 10, or 80\%) or in one of the related categories (20\%). \ \
(b)  Precision at $k$ 
when the distillation step is omitted}}
\end{figure*}

We
now
examine the 150 items that were most highly ranked by our system,
which were 
subsequently
annotated as just described.
(For full system output that includes the unannotated items, see \url{http://www.cs.cornell.edu/~cristian}.  We would welcome external annotation help.)  
As shown in Figure \ref{fig:bars}a,
which depicts precision at $k$ for
various values of $k$, our system performs very well.  In fact, {\em
100\%} of the first 
60 outputs 
are \dm, broadly construed.  It
is also interesting to note the increasing presence of instances that
the judges found hard to categorize as we move further down the
ranking.

Of our 
73 \distractors,
46\% were judged to be members of one of our goal categories.  The
fact that this percentage is substantially lower than our algorithm's
precision at 
both 
73 
and 150 (the largest $k$ we considered) confirms that our judges were not making
random decisions. 
(We expect the
percentage of \pos among the distractors  to be much higher than 0 because
they
were chosen to be similar to our
system's outputs, and so can be expected to also be \pos some fraction
of the time.)

Table \ref{tab:pos} shows the lemmas of 
just the 
\nddm\ {\oplong}s that our algorithm
placed in its top 150 outputs.\footnote{By listing lemmas, we omit variants of
the same word, such as \ex{doubting} and \ex{doubted}, to enhance
readability.  We omit superlatives, comparatives, and conditionals
for brevity.}  
Most of these lemmas are new discoveries, in the sense of not appearing in Ladusaw's \citeyearpar{Ladusaw:80a} (implicit) enumeration of \pos.  
Moreover, the lists of \nddm\ \opslong  that
are used by 
textual-entailment systems are significantly
smaller than that depicted in Table \ref{tab:pos}; for example, \citet{MacCartney+Manning:08a} use only about a dozen (personal communication).

Table \ref{tab:nots} shows examples of the words in our system's top
150 outputs that are either 
clear mistakes or hard to evaluate.
Some of these are due to idiosyncrasies of newswire text. 
For
instance, we often see phrases like \ex{biggest one-day drop in ...},
where \ex{one-day} 
piggybacks on superlatives, and \ex{vowed} piggybacks on the \po \ex{veto}, as in the phrase \ex{vowed to veto}.
\newcommand{\lad}[1]{#1 ({\bf L})}
\newcommand{\ladnote}[2]{\lad{#1}}
\newcommand{\bang}{$\bullet$}
\begin{table*}[htp]
\begin{tabular}{|c|}\hline
\begin{minipage}[t]{6.35in}
\begin{scriptsize}
\begin{multicols}{6}
absence of \\
absent from \\
anxious about \bang\\
\lad{to avoid}\\
to bar\\
barely\\
to block\\
\ladnote{cannot}{not}\\
compensate for \bang\\
to decline\\
to defer\\
\lad{to deny}\\
to deter\\
to discourage\\
to dismiss\\
\lad{to doubt}\\
to eliminate\\
essential for \bang\\
to exclude\\
\lad{to fail}\\
\ladnote{hardly}{hardly ever}\\
to lack\\
innocent of \bang\\ 
to minimize \bang\\
\lad{never}\\
nobody\\ 
nothing\\
to oppose\\
to postpone \bang\\
to preclude\\
premature to\\
to prevent\\
to prohibit\\
\lad{rarely}\\
to refrain from\\
\lad{to refuse}\\
regardless \bang\\
to reject\\
\lad{reluctant to}\\
to resist\\
to rule out\\
skeptical \bang\\ 
to suspend\\
to thwart\\
unable to\\
unaware of\\
unclear on\\ 
unlike\\
\lad{unlikely}\\
unwilling to\\
to veto\\
wary of\\
\ladnote{warned that}{warned against}\\
whenever\\
withstand\\

\end{multicols}
\end{scriptsize}
\end{minipage}
\\\hline
\end{tabular}
\caption{\label{tab:pos}
The 55 lemmas for the 90  \nddmlong \opslong among our algorithm's top 150 outputs. 
\lad{~} marks instances from Ladusaw's list.
\bang ~
marks some of the more interesting cases.
We have added function words 
(e.g., \ex{to}, \ex{for}) to indicate parts of speech or subcategorization; our algorithm does not discover multi-word phrases.
}
\end{table*}

\begin{table*}
 \figtext{ 
\begin{scriptsize}
\begin{center}
\begin{tabular}{|l l l|}\hline 
\textbf{Original} &\textcolor{white}{$\rightarrow$}&\textbf{Restriction
}  \\\hline
Dan is \underline{unlikely} to sing. &{$\stackrel{~}{\Longrightarrow\hspace{-.06in}} \atop \stackrel{\Longleftarrow\hspace{-.08in}/}{~}$}&  Dan is \underline{unlikely} to sing loudly. \\\hline
Olivia \underline{compensates for} eating by exercising. & {$\stackrel{~}{\Longrightarrow\hspace{-.08in}} \atop \stackrel{\Longleftarrow\hspace{-.08in}/}{~}$}& Olivia \underline{compensates for} eating late by exercising. \\\hline  
Ursula \underline{refused} to sing or dance. &{$\stackrel{~}{\Longrightarrow\hspace{-.06in}} \atop \stackrel{\Longleftarrow\hspace{-.08in}/}{~}$}&  Ursula \underline{refused} to sing.\\\hline
Bob would \underline{postpone} singing. & {$\stackrel{~}{\Longrightarrow\hspace{-.06in}} \atop \stackrel{\Longleftarrow\hspace{-.08in}/}{~}$}&  Bob would \underline{postpone} singing loudly.\\\hline 
Talent is \underline{essential for} singing. & {$\stackrel{~}{\Longrightarrow\hspace{-.06in}} \atop \stackrel{\Longleftarrow\hspace{-.08in}/}{~}$}&  Talent is \underline{essential for} singing a ballad.\\\hline 
She will finish \underline{regardless of} threats. & {$\stackrel{~}{\Longrightarrow\hspace{-.06in}} \atop \stackrel{\Longleftarrow\hspace{-.08in}/}{~}$}& She will finish \underline{regardless of}  threats to my career.\\\hline
\end{tabular}
\end{center}
\end{scriptsize}
}
\caption{
Example demonstrations that the underlined expressions (selected from Table \ref{tab:pos}) are \pos: the sentences on the left entail 
those
on the right.   We also have provided $\Longleftarrow\hspace{-.18in}/$\ \ \ indicators because the reader might find it helpful to reason in the opposite direction and see that these expressions are not \umlong.
}
\end{table*}

\begin{table}
\begin{center}
\figtext{ 
\begin{scriptsize}
\begin{tabular}{|lll|l|}\hline
& \textbf{\no} & & \textbf{Hard} \\\hline
almost & firmly & one-day & approve\\ 
ambitious & fined & signal & cautioned \\
considers & liable & remove & dismissed \\
detect & notify & vowed & fend \\
\hline 
\end{tabular}
\end{scriptsize}
}
\end{center}
\caption{\label{tab:nots} 
Examples of words judged to be either not in one
of our monotonicity categories 
of interest 
(\no)
or hard to evaluate
(Hard).  
}
\end{table}

\paragraph{Effect of distillation}
In order to evaluate the importance of the \distillation process, we 
study how the results change when \distillation is omitted
(thus using 
as 
score function 
$\scorefn$ from Equation \ref{un_distilled_score} rather than $\distscorefn$).
 When comparing the results (summarized in Figure \ref{fig:bars}b) with those of the complete system (Figure \ref{fig:bars}a) we observe that the \distillation
indeed has 
the desired effect: the number
of
 highly ranked words 
that 
are annotated as \no
decreases after distillation.  This results in an increase of the precision at $k$ 
ranging from 5\% to 10\% (depending on $k$), as
can be observed by comparing the 
height of the composite bars in the two figures.\footnote{The words annotated
``hard''
do not affect this increase in precision.
 }
 
 Importantly, this improvement does indeed seem to stem at least in part from the distillation process handling the piggybacking problem.  To give just a few examples: 
\ex{vigorously} is pushed down from rank 48 (undistilled scoring) to rank 126 (distilled scoring), \ex{one-day} from $25^{\rm th}$ to $65^{\rm th}$, \ex{vowed} from $45^{\rm th}$ to $75^{\rm th}$, and \ex{Milken} from $121^{\rm st}$ to $350^{\rm th}$.

\subsection{Other Results}

It is natural to ask whether the (expected) decrease in precision at
$k$ is due to the algorithm 
assigning
relatively low scores to \pos, so
that they do not appear in the top 150, or 
due to there being no more
more true \pos to rank.  We cannot directly evaluate our method's recall because no comprehensive list of \pos exists.  However, to get a rough impression, we can check how our system ranks the items in the largest list we are aware of, namely, the Ladusaw (implicit) list mentioned above.  
Of the 
31
\po lemmas on this list
(not including the 10 well-known \pos), 
only 7 of those frequent enough to be considered by our algorithm are not in its top 150 outputs, and only 5 are not in the top 
300.  Remember that we only annotated the top 150 outputs; so, there may be many other \pos between positions 150 and 
300.

Another way of evaluating our method would be to assess the effect of our newly discovered \pos on downstream RTE system performance.  
There are two factors to take into account.   First, 
the \pos we discovered are quite prevalent in naturally occurring text\footnote{However, RTE competitions do not happen to currently stress inferences involving monotonicity.  The reasons why are beyond the scope of this paper.
} : the 
90 \nddm\ \opslong appearing in our algorithm's top 150 outputs occur in 
111,456 sentences in the BLLIP corpus
(i.e., in 6\% of its sentences).
Second, as previously mentioned, systems do already account for monotonicity to some extent --- but they are limited by the fact that their \po lexicons are restricted mostly to well-known instances; to take a concrete example with a publicly available RTE system: Nutcracker \citep{Bos+Markert:06a}  correctly infers that \ex{We did \underline{not} know the disease spread} entails \ex{We did \underline{not} know the disease spread quickly} but it fails to infer that  \ex{We \underline{doubt} the disease spread} entails \ex{We \underline{doubt} the disease spread quickly}.  So, systems can use monotonicity information but currently do not have enough of it; our method can provide them with this information, enabling them to handle a greater fraction of the large number of naturally occurring instances of this phenomenon  than ever before.

\section{Related work not already discussed}
\citet{Magnini:08a}, in describing modular approaches to textual
entailment, hints that NPIs may be used
within a negation-detection sub-component.

There is a substantial body of work in the linguistics literature
regarding the definition and nature of polarity items
\citep{pi-bib:08a}.  
However, very little of this work is
computational.  
There has been passing speculation that one might want
to learn polarity-inverting verbs \citep[pg. 47]{Christodoulopoulos:08a}.
There have also been a few projects on the discovery of NPIs, which
is the converse of the problem we consider.
\citet{Hoeksema:97a} discusses some of the difficulties with
corpus-based determination of NPIs, including ``rampant'' polysemy and
the problem of ``how to determine independently which predicates
should count as negative'' --- a problem which our work addresses.
Lichte and Soehn \citep{Lichte:05a,Lichte+Soehn:07a} consider 
finding German NPIs using a method conceptually similar in some
respects to our own, although again, 
their objective is the reverse of ours.
Their discovery statistic for single-word 
NPIs is the ratio of 
within-licenser-clause occurrences to total occurrences, where, to
enhance precision, the
list of licensers was filtered down to a set of
fairly unambiguous, 
easily-identified items.  
They do not consider 
\distillation, which we found to be an important
component of our \podash-detection algorithm.  Their evaluation scheme,
unlike ours, did not employ a bias-compensation mechanism.  They did
employ a collocation-detection technique to extend their list to
multi-word NPIs, but our independent experiments with a similar technique (not
reported here) did not
yield good results.

\section{Conclusions and future work}

To our knowledge, this work represents the
first attempt to 
discover 
\dmlong\ {\oplong}s.
We introduced a 
unsupervised algorithm
that is
motivated by research in
linguistics but 
employs
simple distributional statistics in a novel
fashion.
Our algorithm
is highly accurate and 
discovers many reasonable \pos that are missing from 
pre-existing 
manually-built
lists.

Since the algorithm is 
resource-lean ---  requiring no parser or tagger
but only a list of NPIs --- it can be immediately applied to 
languages
where such lists exist, such as German and Romanian
\citep{Trawinski+Soehn:08a}. On the other hand, although the results
are already quite good for English, it would be interesting to see
what improvements could be gained by using more sophisticated
syntactic information.

For languages where NPI lists are not extensive,
one could envision applying
an iterative co-learning approach:
use
the newly-derived \pos to infer 
new
NPIs, and 
then discover even more new \pos given the new NPI list.
(For English, our initial attempts at bootstrapping from our initial NPI list on the BLLIP corpus  did not lead to substantially improved results.) 

In practice, 
subcategorization is an important feature to capture.  
In Table
\ref{tab:pos}, we indicate which subcategorizations
are
\dm.  An interesting extension of our work would be to try
to automatically distinguish particular \dm subcategorizations that
are lexically 
apparent, e.g., \ex{innocent} (not \dm) vs. \ex{innocent of} (as in \ex{innocent of burglary}, \dm).

Our project provides a connection (among many) between the creation of textual
entailment systems (the domain of language engineers) and the
characterization of \pos
(the subject of study and debate among linguists).
The prospect that our method might potentially
eventually be refined in such a way so as to shed at least a little
light on linguistic questions is a very appealing one, although we
cannot be certain that any progress will be made on that front.

\setstretch{0.8}
{\footnotesize
\paragraph{Acknowledgments}  
We thank Roy Bar-Haim, Cleo Condoravdi, and
Bill MacCartney for sharing their systems' lists and
information about their work with us; Mats Rooth for helpful conversations; 
Alex Niculescu-Mizil for technical assistance; 
and Eugene Charniak for reassuring remarks.
We also thank 
Marisa
Ferrara
Boston, Claire Cardie, Zhong Chen, Yejin Choi, Effi Georgala,
Myle Ott, Stephen Purpura, and Ainur Yessenalina at Cornell
University, 
the 
UT-Austin
NLP group,
Roy Bar-Haim,
Bill MacCartney, 
and  the anonymous reviewers for 
for their comments on this paper.
This paper is based upon work supported in part by 
DHS grant N0014-07-1-0152,
National Science Foundation
grant No.
 BCS-0537606,  
a Yahoo!\ Research Alliance gift, 
a CU Provost's Award for Distinguished Scholarship, and a CU Institute for the Social Sciences Faculty Fellowship.
Any opinions, findings, and conclusions or recommendations expressed are those
of the authors and do not necessarily reflect the views or official policies,
either expressed or implied, of any sponsoring institutions, the U.S.\
government, or any other entity.
} % close footnotesize
\small
\bibliographystyle{plainnat}

\end{document}